# Guess-And-Verify Heuristics for Reducing Uncertainties in Expert Classification Systems

**Yuping Qiu**
U S WEST Advanced Technologies
4001 Discovery Drive
Boulder, CO 80303

**Louis Anthony Cox, Jr.**
U S WEST Advanced Technologies
4001 Discovery Drive
Boulder, CO 80303

**Lawrence Davis**
Tica Associates
36 Hampshire Street
Cambridge, MA 02139

## Abstract

An expert classification system having statistical information about the prior probabilities of the different classes should be able to use this knowledge to reduce the amount of additional information that it must collect, e.g., through questions, in order to make a correct classification. This paper examines how best to use such prior information and additional information-collection opportunities to reduce uncertainty about the class to which a case belongs, thus minimizing the average cost or effort required to correctly classify new cases.

## INTRODUCTION

Practitioners often wish that their AI consulting systems would display not only mechanical rationality -- e.g., exhibited in systematic, goal-directed questioning leading to sound conclusions justified by explicit rules -- but also inspiration, indicated by an ability to quickly guess correct answers based on minimal information. This paper presents a simple model of an expert system task, namely, classification of cases based on information elicited interactively from the user. It discusses heuristics for using prior statistical information to guess the correct class of a case as quickly (or, more generally, as cheaply) as possible. Motivating applications for such heuristics could include the following:

o *Design of interactive dialogues for diagnostic and advisory systems:* In many applications of expert consultation systems, the user and the system should share a goal of minimizing the effort that the user must supply to get a query answered. The user's effort is expended in providing answers to questions asked by the system. The system determines which questions to ask based on its knowledge base and on some reasoning about what conjunctions of facts will suffice to justify conclusions. This paper presents heuristics for using statistical information about the prior probabilities that different conclusions are correct to adaptively formulate a sequence of questions that will minimize the expected effort required from the user.

o *Pattern recognition and string-matching:* Numerous applications in telecommunications, molecular biology, $C^3I$ and computer science require a system to repeatedly identify patterns and/or classify strings of incoming symbols. Our heuristics provide guidance on which symbols to examine first to match a new string to one of a finite number of known possible ones as quickly or cheaply as possible.

o *Optimizing applied scientific research:* Our heuristics can also be used to design cost-effective applied research programs -- e.g., to determine whether a chemical is a probable human carcinogen -- by adaptively sequencing the tests to be performed. Performing a diagnostic test can be viewed as a form of question-answering, and therefore falls within the scope of a theory for minimizing the expected cost of questions that must be answered in order to reach a conclusion.

o *Optimal diagnosis:* If a complex system has a known set of possible failure states with known prior probabilities, then the problem of sequentially inspecting components to minimize the expected cost of determining which failure state the system is in is an instance of the problem solved in this paper.

## 1. PROBLEM FORMULATION

A *logical classification expert system* for assigning cases to classes (where classes may be interpreted as diagnoses, predictions, or prescriptions, for example) consists of a set of rules that determine the unique class to which each case belongs from the truth values of various propositions about the case. ("Unknown" can be included as one of the classes.) Suppose that the propositions that determine class membership have been reduced to disjunctive normal form (DNF); then the classification expert system can be expressed as a set of rules, each of the form

(case z belongs to class j) if

[propositions $(X_{j1}, ..., X_{jn})$ all hold for case z]     (1).

Here, the $X_{ij}$ are atomic propositions that can be asserted about the case (they are literals or negations of literals, so that the rules need not be strictly Horn); the disjunctions of the DNF have been distributed across multiple rules; and the conjunctive terms have become the antecedent conjunctions of the rules. (The dummy variable z, which might be regarded as implicitly universally quantified over the set of all cases, is bound in any application, so that propositional rather than predicate calculus applies.)

An expert classification system whose rules are in the canonical form (1) can be represented by an M x N matrix of ones, zeros, and blanks and an M-vector of class labels, where M ≥ the number of distinct classes and N is the total number of literals that can be asserted (or denied) about a case. The interpretation is that columns correspond to the literals, rows to the classes, and if the pattern of truth values of the literals for a case matches the pattern of zeros and ones in row i (with the usual coding 0 = false, 1 = true), then the case belongs to the class whose label is associated with row i. The same label could appear more than once in the M-vector of labels, since there may be more than one way to prove membership in a class. However, this paper assumes that there is a one-to-one correspondence between classes and rows. This is a natural restriction in many cases (see below). Blanks in the matrix may be used to represent "don't care" conditions, i.e., values of literals that do not enter into the antecedent of a rule. Such a matrix may be called the *canonical matrix* for a logical classification system. It maps each N-vector of truth values for literals into exactly one corresponding class label. Let L ≤ M denote the number of distinct classes: then the classification system may be viewed abstractly as computing a discrete classification function f: $\{0, 1\}^N \rightarrow \{1, 2, ..., L\}$. The challenge is to develop algorithms or heuristics for computing the value of this function, for any set of instance data, as quickly (or as inexpensively) as possible.

Any set of propositional logic rules that maps measurements (truth values of literals representing observable quantities) into classes can be represented in the canonical form. Thus, complex rules such as the following may be included in the original rule base: "((a case belongs to either class 1 or 2) if ((attribute A is present and attribute B is absent) unless (the case has property D but not property E))." Reducing such arbitrary rule bases to a minimal DNF requires exponential time in general. However, in practice, few classification rule bases need or use the full flexibility of propositional logic. Instead, classification is typically based on direct inspection of which attributes are present or absent in a case. The remainder of this paper examines classification problems in which information about a case is obtained via a set of (in general costly) tests for properties that may be present or absent; and the pattern of properties for a case is drawn from a small finite set of distinct possible patterns. Thus, each class corresponds to exactly one row in the canonical matrix (i.e., a unique set of truth values) and there is a one-to-one correspondence between classes and rows. No further "pre-processing" to obtain a DNF form is assumed. The costly tests will be referred to generically as "inspections": they may represent questions put to the user, retrievals from a remote database, laboratory assays, measurements, etc., depending on the application. This class of problems is interesting for practical problems and our heuristics for solving it suggest extensions to the more general case of non-unique configurations of truth values for classes.

Let x denote the (initially unknown) vector of truth values for the set of N literals applied to a new (as-yet uninspected) case. (In pattern recognition applications, x could be a binary "feature vector" stating which features are present and which ones absent.) Let A be the canonical matrix of an expert classification system. If x were known, it could be compared to the rows of A until a match is found (i.e., until the truth values of the components of x match the truth values specified in the row.) Then the classification of x would be given by the class label of that row. In practice, the following two considerations make classification less straightforward:

o *It may be costly to observe the components of x.* Determining whether a proposition holds (e.g., whether a feature is present) may require the user to perform an expensive test. If the classification system obtains information about a specific case by questioning the user during the course of a consultation session, for example, then each question imposes some burden on the user even if no experiment is required to obtain the answer. To capture this pragmatic aspect of classification, we will let c denote an N-vector of *inspection costs* for determining the truth values of the N components of x.

o *The answers to some questions may be more or less predictable in advance.* For example, after classifying many cases, the relative frequencies of different class memberships may prove to be stable and predictable. Such statistical information should be exploited by the system. Intuitively, relevant, reliable statistical information can be used to help shift the interactive questioning and classification process away from a purely systematic search toward a more efficient "guess and verify" strategy. To formally model the optimal use of statistical information to reduce the average amount of effort or cost needed to find the right answer, we will henceforth assume that there is given an L-vector of prior *class probabilities,* denoted by p.

In summary, a general classification problem with prior information about class probabilities is represented in canonical form by a triple (A, p, c) and a vector of uncertain but discoverable truth values, x, where A is the canonical matrix; x is one of L possible truth value configurations (typically interpreted as patterns or "states" for a case) having respective probabilities of p(1), p(2), ..., p(L) for classes 1, 2, ..., L; and c is the vector of inspection costs, with $c_i$ being the cost of determining the truth value of literal i for i = 1, 2, ..., N. A different probability model results if it were assumed that the truth values of the N literals are statistically independent random values (e.g., the literals could describe the states, working or not, of each of several independent components of a complex reliability systems.) Heuristics for finding near-optimal questioning sequences have been discussed in detail for this case in (Cox 1990). This paper examines the analogous problem for the case where class probabilities, rather than truth value probabilities for literals, are known.

A candidate *solution* for a logical classification problem (A, p, c) can be represented as a rooted binary classification tree in which each nonterminal node represents a literal (a property in our simplified model), the two branches descending from such a node correspond to the two possible truth values or test outcomes for that property (left branch = absent, right branch = present), and the terminal nodes contain the names of the classes. Such a tree specifies which property the system should ask about next, given the answers it has received so far, until it has obtained enough information to deduce which class the case belongs to. (A slight generalization would allow it to stop short of establishing the correct class with certainty, instead halting and making a best guess when the expected value of additional information, measured in terms of the expected reduction in the cost of classification error, is less than the expected cost to collect it. The algorithms discussed in this paper can easily be modified to incorporate this refinement.) The *expected cost* of a candidate solution can be defined recursively as follows:

(i) The expected cost of a terminal node is zero.

(ii) The expected cost of a nonterminal node is the inspection cost for the corresponding literal (i.e., for testing whether that property holds) plus the sum of the costs of each of its two child nodes weighted by their probabilities (i.e., by the probability that the case has the property for the right child node, and weighted by one minus this probability for the left child node.)

An *optimal solution* is a classification tree having minimum expected cost among all possible classification trees. It thus represents a cost-effective strategy for reducing uncertainty about class membership by optimally sequencing questions about a case's properties. Section 2 presents algorithms and heuristics for constructing optimal or nearly optimal classification trees.

As an example, consider an A matrix consisting of the following four pattern vectors (rows): (1, 0, 0, 1), (0, 1, 0, 1), (1, 0, 1, 0), and (0, 1, 1, 1). Let the prior probabilities of these four patterns be 0.4, 0.2, 0.3, and 0.1, respectively; thus, p = (0.4, 0.2, 0.3, 0.1) The inspection cost vector is c = (3, 1, 4, 1). Given a randomly generated pattern instance $x = (x_1, x_2, x_3, x_4)$, what is the best inspection strategy for examining the components of x to determine which pattern it is an instance of? The answer is not obvious to a human even though the problem is small. Using the dynamic programming procedure discussed in Section 2, an optimal answer is found to be as follows (see Figure 1): Inspect component 2 of x first (corresponding to column 2 of the A matrix.). If there is a zero in column 2 (which occurs with probability 0.4 + 0.3 = 0.7) then inspect component 4 next; otherwise, inspect component 3 next. The outcome of these two inspections will determine which pattern x is an instance of, and no other inspection strategy has a smaller expected cost. (The expected cost of this strategy is 2.9. There is another optimal strategy that also achieves this expected cost that the reader may try to find.) The CPU time required to find this solution via dynamic programming is less than 0.01 seconds.

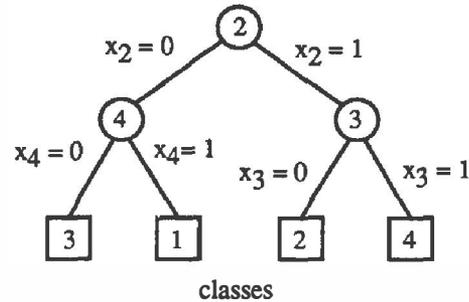

Figure 1. An example classification tree.

## 2. EXACT AND HEURISTIC PROCEDURES

### 2.1 Exact Solutions

Constructing an optimal (minimum expected-cost) classification trees is intrinsically difficult. Although exact optimal solutions can be found using dynamic programming, the time-complexity required is in general exponential in the number N of columns of A. After stating these results, this section introduces two heuristics based on the ideas of information-theoretic efficiency and searching for least-cost distinguishing subsequences ("signatures") for the different classes. These heuristics have complementary strengths, allowing them to be hybridized to create very effective approximate procedures.

*Theorem 1:* *The problem of finding optimal (minimum expected-cost) classification trees for systems (A, p, c) is NP-hard.*

*Proof:* The proof is based on a polynomial-time reduction of the well-known NP-hard *set covering problem* to an optimal classification problem with equal costs. It is given in the appendix. Notice that if an "inspection" is interpreted as a data base retrieval of a stored value, then this result is compatible with the major result in (Greiner 1991).

Optimal classification strategies for small problems (e.g., for A of dimension less than about 20 x 20) can be found by dynamic programming, as follows. Let C(A, p, c) denote the minimum expected cost of classifying a case using system (A, p, c). Then an optimal classification tree can be constructed by solving the following dynamic programming recursion for j, the index of the first column of A to be inspected:

$$C(A, p, c) = \min_{1 \le j \le n} [c_j + w_{j0}C(A_{j0}, p_{j0}, c) + (1 - w_{j0})C(A_{j1}, p_{j1}, c)] \quad (2)$$

where $A_{jk}$ denotes the submatrix consisting of those rows of A having a truth value of k in column j, for k = 0, 1; $p_{jk}$ = the normalized probability vector for the submatrix $A_{jk}$, found by rescaling the row probabilities for $A_{jk}$ to sum to 1; and $w_{jk}$ = prior probability that the truth value in column j is k, for k = 0, 1 (i.e., $w_{jk}$ = the sum of the prior row probabilities for all rows in $A_{jk}$.) The time complexity of this recursion is exponential in N, the number of columns of A (it is N! in the special case where A is an identity matrix and of order not greater than $n(n-1)...(n - m + 2)2^{m-1}$ in general.) If the number of rows of A is bounded by a constant, K, however, then the time complexity is $O(N^{K-1})$, i.e., it is polynomial in N for any given number of rows of A.

## 2.2 Information-Theoretic and Signature-Based Heuristics

When all inspection costs are equal, the optimal classification problem is to minimize the average number of questions (inspections) needed to determine the class of a case. If the rows of A could be designed by the user, then a sharp lower bound on the average number of questions required would be given by the entropy of the class probability vector p. A larger number of questions is typically required when A is exogenously specified, but the criterion of entropy reduction still provides a useful guide for selecting columns to minimize the number of questions that must be asked. If inspections are costly, then this criterion can be generalized to entropy reduction per unit cost.

*Information-theoretic (entropy) heuristic:* At any stage of the sequential inspection process, inspect next the column that yields the greatest expected reduction in classification entropy per unit of inspection cost, given the results of all inspections made so far.

The classification entropy at any stage is a measure of the uncertainty about the true class: it is defined as $H(p) = -\Sigma_i p_i \log_2(p_i) = E[1/\log_2(p_i)]$ where the sum is taken over all classes i that are consistent with the observations made so far and where the prior probabilities $p_i$ for this subset of classes have been normalized (rescaled) to sum to 1. Intuitively, the information-theoretic heuristic always prescribes performing next the test that is expected to be "most informative" per unit cost, where informativeness is defined by expected entropy reduction. Letting H(A) denote the classification entropy of matrix A, this amounts to always picking next the column j that minimizes $[w_{j0}H(A_{j0}) + (1 - w_{j0})H(A_{j1})]/c_j$. Motivation for the information-theoretic heuristic is provided by the following often rediscovered "folk theorem":

*Lemma:* Let T be a classification tree for matrix A. Define the <u>weight</u> of node j in T as the probability of the subtree rooted at node j (i.e., it is the sum of the probabilities of the classes at the terminal nodes in this tree.) Define the weight of the whole tree T as the sum of the weights of its nodes. If the prior class probabilities are equal (uniform probabilities, each equal to 1/M) and inspection costs are uniform (all components of c are equal) then T is optimal if and only if it has minimum weight among all classification trees. More generally, for uniform costs and arbitrary prior class probabilities, the information-theoretic heuristic is equivalent to choosing columns in order to balance, as nearly as possible, the probabilities of the right and left subtrees at each node.

In the special case of uniform p and c, the maximum-entropy column is the one with the most nearly equal numbers of ones and zeros among the remaining (unexcluded) rows of A, and the information-theoretic heuristic calls for inspecting next this most nearly balanced column. The lemma suggests that a balanced tree tends to be preferred to an unbalanced one. The heuristic is not always optimal even when p and c are uniform, however, since myopic optimization (always making the one-step most informative inspection) does not necessarily lead to global optimization (a multi-step most informative sequence of inspections.)

Similar entropy-reduction heuristics without consideration of costs have been widely applied in the machine learning literature on *recursive partitioning algorithms* for learning classification rules from noisy data (Breiman et al 1984; Quinlan 1986). Our optimal classification problem is different, since (A, p, c) contains an explicit model of the data-generating process (making it unnecessary to learn it from sample data) and since the classification rules are already known. The entropy reduction criterion finds a new use in this setting by suggesting how to use the rules most cost-effectively to make a classification decision.

An alternative approach to cost-effective information collection follows a *guess-and-verify strategy*. Consider first the case of uniform inspection costs. Rather than looking for the most informative column to inspect, the guess-and-verify approach instead identifies the row of A that is most likely to be the true one, given all observations made so far. Then, it tries to prove or disprove efficiently the hypothesis that the case being classified matches this most likely row. To test this hypothesis cost-effectively, a *signature* is created for each row. A signature for row i is a subset of column values such that this combination of column values occurs only in row i. Thus, if these values are observed for a case, then the case belongs to class i. Given a signature for the currently most likely row, the next column to be inspected is the one corresponding to the element of its signature that is *least* likely to match observations (If the case being classified does not have the hypothesized signature, then it is desirable to find this out as quickly as possible instead of wasting resources making easy matches, only to eventually discover that the hypothesized signature is wrong after all.)

In the general case of unequal inspection costs, the guess-and-verify heuristic works as follows.

*Signature Heuristic:*

*Step 1 (Signature Generation):* Generate a signature for each row using the following greedy heuristic (Liepens and Potter 1991). For each row, i, choose some initial column, $j_0$, as a seed. The value in column $j_0$ of row i discriminates this row from all rows with the opposite value in column $j_0$ (assuming that any informationless column, having all identical entries, has been deleted.) Add to this partial signature $\{j_0\}$ the column having the greatest ratio of discrimination probability to inspection cost, where the "discrimination probability" for a column is defined as the sum of the probabilities of the rows that differ from row i in this column. Continue to select columns by this ratio criterion and add them to the partial signature until a complete signature (discriminating row i from all other rows) has been formed. Apply this signature-growing procedure starting with each of the N possible elements of row i as a seed, and choose the signature with the smallest total cost (sum of the inspection costs of its columns) as the one to be used for row i.

*Step 2 (Row Selection):* At any stage, choose as the next row to test the one with the greatest ratio of success probability to cost. The "success probability" in the numerator is the conditional probability (after eliminating all rows that are inconsistent with previous observations) that the case is an instance of this row. The "cost" in the denominator is the cost of the signature for the row, i.e., it is the sum of the inspection costs of its as-yet unobserved components.

*Step 3: (Column Selection):* The next column to be inspected is the one that contains the *least* likely element in the signature of the selected row.

Interpretively, Step 2 selects the "easiest" problem to work on (as measured by the probability-to-cost ratio) while Step 3 prescribes working on the "hardest" part of this problem first, so as not to waste effort prematurely on the easier parts. The "problems" considered consist of proving that a case instantiates a particular row by inspecting its components.

## 3. EXPERIMENTAL RESULTS AND A HYBRID HEURISTIC

To evaluate the performance of the information-theoretic and signature heuristics, we randomly generated several thousand test problems and solved them exactly, using the dynamic programming recursion in equation (2), and also approximately using the heuristics. Although careful discussion of the experimental design used and the results obtained are beyond the scope of this paper, the highlights of the experiments were as follows.

1. For randomly generated A matrices of a given size (numbers of rows and columns), the comparative performances of the information-theoretic and signature heuristics depend systematically on the entropy of the prior class probability vector, H(p), and on the coefficient of variation of the inspection costs (i.e., on the ratio of the standard deviation of the components of c to their mean value.) Therefore, for experiments with given numbers of rows and columns of A, we generated a fixed number (50 in exploratory experiments and up to three hundred in final experiments) of classification problems in each cell of a 5 x 10 grid of *entropy x coefficient of cost variation* values. The five entropy "bins" used for class probability vectors were chosen to cover the range from zero to the maximum possible entropy (equal to $\log_2 M$ for a problem with M rows) in equal intervals on a log scale; thus, for example, for a problem with M = 10 rows as possible patterns, we would generate an equal number of prior class probability vectors in each of the entropy ranges 0.0 - 0.66, 0.67 - 1.3, 1.3 - 2.0, 2.0 - 2.65, 2.65 - 3.3. The inspection cost vectors were binned into ten intervals based on their coefficients of variation, ranging systematically from 0.0 - 0.1 through 0.9 - 1.0. The distribution of relative errors (compared to the exact solution) was studied for each heuristic in each of the 50 cells in this grid, for a variety of problem sizes.

Tables 1-3 summarize the results of these experiments. (To save space, only five of the ten coefficient-of-variation columns are shown.) These results are for A matrices of dimension 10 x 10, i.e., ten patterns and ten properties. The numbers shown in each cell are the average relative

**Table 1: Relative Percentage Errors for the Entropy Heuristic (10 x 10 Problems)**

| | *Coefficients of Variation of Costs* | | | | |
|---|---|---|---|---|---|
| Entropy | 0.1 | 0.3 | 0.5 | 0.7 | 0.9 |
| 0 - 0.67 | 20.1 | 17.1 | 15.4 | 13.4 | 8.6 |
| 0.67 - 1.3 | 13.8 | 16.3 | 10.8 | 8.0 | 7.8 |
| 1.31 - 2.0 | 12.75 | 10.8 | 7.3 | 7.4 | 5.6 |
| 2.1 - 2.65 | 6.2 | 5.1 | 4.5 | 3.8 | 3.9 |
| 2.66 - 3.3 | 1.8 | 1.8 | 1.5 | 2.3 | 1.6 |

**Table 2: Relative Percentage Errors for the Signature Heuristic**

| | *Coefficients of Variation of Costs* | | | | |
|---|---|---|---|---|---|
| Entropy | 0.1 | 0.3 | 0.5 | 0.7 | 0.9 |
| 0 - 0.67 | 0.39 | 0.58 | 1.5 | 0.56 | 0.83 |
| 0.67 - 1.3 | 8.2 | 7.5 | 5.0 | 6.2 | 4.3 |
| 1.31 - 2.0 | 9.6 | 9.4 | 7.9 | 6.9 | 7.2 |
| 2.1 - 2.65 | 8.2 | 10.4 | 8.6 | 8.2 | 5.4 |
| 2.66 - 3.3 | 7.9 | 10.8 | 9.2 | 9.6 | 5.9 |

**Table 3: Relative Percentage Errors for the Hybrid (Entropy/Signature) Heuristic**

| | *Coefficients of Variation of Costs* | | | | |
|---|---|---|---|---|---|
| Entropy | 0.1 | 0.3 | 0.5 | 0.7 | 0.9 |
| 0 - 0.67 | 0.14 | 0.11 | 0.81 | 0.15 | 0.27 |
| 0.67 - 1.3 | 3.9 | 2.9 | 1.5 | 1.4 | 1.35 |
| 1.31 - 2.0 | 3.2 | 2.7 | 2.1 | 1.35 | 1.6 |
| 2.1 - 2.65 | 1.8 | 1.6 | 1.3 | 0.44 | 1.65 |
| 2.66 - 3.3 | 0.48 | 0.65 | 1.08 | 0.98 | 0.64 |

errors for 50 randomly generated test problems falling in the corresponding entropy bin for p and coefficient of variation bin for c.

2. From the experimental results in Tables 1-3, it is clear that the information-theoretic and signature classification heuristics have complementary strengths. The signature algorithm does better (by up to 30-fold on 10 x 10 problems) on relatively uniform problems having low entropy and low coefficient of cost variation. The information-theoretic algorithm does better (typically by a factor of between 2 and 9 on 10 x 10 problems) on all problems with sufficiently high entropy of the prior class probability vector; at lower entropy levels, its performance (measured by relative error compared to the true optimum) improves as the coefficient of variation of the cost vector increases. For example, on 10 x 10 problems, the signature algorithm dominates in the first two (low) entropy bins (i.e., for problems with prior classification entropies of less than 1.4 bits) while the information-theoretic algorithm dominates in the upper two entropy bins (problems with prior classification entropies of greater than 2.6 bits.)

3. The average performance of the heuristics deteriorates gradually as problem size increases. We examined performance on problems ranging from 4 x 4 to 16 x 16, with a variety of non-square matrices for comparison. (For problems of size 16 x 16, the dynamic programming algorithm took on the order of 1 hour on a Symbolics 3630 lisp machine to find optimal solution trees, and we have not yet considered larger problems.) For example, average percentage errors for the signature algorithm increased from about 1.4% on a set of one hundred 5 x 5 test problems to about 8.7% on a set of one hundred 14 x 14 test problems.

4. Because the information-theoretic and signature algorithms have complementary strengths, a hybrid of the two works much better than either of them alone. They can be hybridized by generating subtrees from each node using both heuristics and keeping the one producing the better result (lower expected cost.) Using such a hybrid heuristic, we were able to reduce the average error on the test set of 10 x 10 problems from about 9% to about 1% (see Table 3.) Although the theoretical worst-case computational complexity of this hybrid heuristic is exponential, in practice it appears to take less than twice as long to run as the signature heuristic alone.

5. Relative computational efficiency for the hybrid heuristic and both of its components (the information-theoretic and signature heuristics) increase with problem size. For problems smaller than 8 x 8, the dynamic programming solution takes less than a second (even without correcting for garbage collection and other sources of machine-induced variation) and there is little point in using a heuristic. For 14 x 14 problems, the dynamic programming solution takes over ten minutes, while the most computationally expensive version of the hybrid (trying both signature and information-theoretic criteria at each node) takes about 2 seconds. Computing time for the dynamic programming algorithm rises sharply with problem size: 13 x 13 problems take less than six minutes to solve, while 16 x 16 problems take over an hour. The two heuristics take less than a second for all problem sizes examined, with the hybrid taking slightly longer (about 1.5 seconds on 10 x 10 problems compared to 0.92 seconds for the signature heuristic and 0.24 seconds for the entropy heuristic.) For larger problems, the hybrid heuristic is expected to achieve better than 95% of the optimal solution in less than 1% of the time needed to reach full optimality. It is easily checked that both our heuristics have polynomial-time complexity (bounded above by $O(mn^2 \cdot \min\{m, n\})$ for the signature heuristic and $O(mn \cdot \min\{m, n\})$ for the entropy heuristic based on the numbers of operations that must be performed. The factor $\min\{m, n\}$ arises from the fact that each inspection eliminates at least one row and one column.) Thus, their advantage relative to the exact solution procedure continues to increase as problem sizes grow larger.

4. CONCLUSIONS

We have presented two heuristics for obtaining approximate solutions to the NP-hard problem of deciding in what order to ask questions so as to minimize the average cost of classifying cases. The preliminary experimental results reported here suggest that these heuristics are effective on small problems. The most exciting discovery, however, is that they can easily be hybridized to obtain substantial improvement in performance. Additional experiments are currently being designed and extensions of these results to arbitrary logical classification systems (when more than one row may represent the same class) are being developed. For practical purposes, it appears that the hybrid of the signature and entropy heuristics achieves near-optimal solutions with at most a few seconds of computational effort in problems with fewer than about 20 classes. (The time required by the heuristics is relatively insensitive to the number of attributes available to be inspected since signature length grows slowly as columns are added.) This range of problem sizes contains many applications of practical interest.

## Appendix: Proof of Theorem 1

*Proof of Theorem 1.* We present a polynomial-time reduction from the set covering problem to a specially constructed pattern classification problem $(A, p, c)$ with equal costs. The set covering problem is a well-known NP-complete problem and can be stated as follows:

*Given a collection $F$ of subsets of a set $S$ and a positive integer $k$, does $F$ contain a subset $F'$ (called a cover) with $|F'| \leq k$ and such that $\{\cup f \mid f \in F'\} = S$ ?*

For an instance of the set covering problem, we construct a corresponding pattern classification instance in polynomial time. Let $m = |S| + 1$ and $n = |F|$. Define the pattern matrix $A$ as follows: $a_{ij} = 1$ if $i < m$ and the ith element of $S$ is in the jth subset in $F$; $a_{ij} = 0$ otherwise. Finally, set $p_i = (m-1)^{-1}(n+1)^{-1}$ ($1 \leq i < m$), $p_m = n/(n+1)$, $c_i = 1$ ($1 \leq i \leq n$).

Given this construction, if the true pattern is $a_m$ in the pattern classification problem, then it can be distinguished from the other patterns by inspecting the components corresponding to an arbitrary cover in the set covering problem. Conversely, the components that must be inspected in order to separate $a_m$ from the other patterns always induce a cover in the set covering problem. An inspection policy is represented by a binary decision tree with $m-1$ nonterminal nodes (decision nodes labeled by components) and $m$ terminal nodes (corresponding to the $m$ patterns). For any given inspection tree, let $E_i$ denote the subset of distinct components that are on the path from the top node to the terminal node corresponding to pattern $i$. Then $|E_i| \leq n$ and $E_m$ corresponds to a cover in the set covering problem. Let $v$ denote the expected inspection cost associated with the inspection tree: then it must be the case that $p_m|E_m| < v = p_1|E_1| + \cdots + p_m|E_m| \leq p_m|E_m| + (1-p_m)n$.

To complete the proof we show that the answer to the set covering problem is "yes" if and only if the optimal inspection tree in the corresponding classification problem satisfies $|E_m| \leq k$. If $|E_m| \leq k$, then the subsets corresponding to the components in $E_m$ forms a desired cover in the set covering problem. On the other hand, if $|E_m| \geq k+1$, then the minimum expected inspection cost satisfies $v^* > p_m(k+1)$. We now show by contradiction that there is no cover with size less than or equal to $k$ in this case. Suppose that, to the contrary, there is a cover $F'$ composed of $k$ or less subsets. Then the components corresponding to $F'$ form the set $E_m$ for some feasible inspection tree. The expected inspection cost of such a tree satisfies $v \leq p_m|E_m| + (1-p_m)n \leq p_mk + (1-p_m)n$. Therefore, we would have $p_m(k+1) < v^* \leq v \leq p_mk + (1-p_m)n$, which implies that $p_m < n/(n+1)$. But this contradicts the fact that $p_m = n/(n+1)$. This completes the proof. ◇